\documentclass[journal,twoside,web]{ieeecolor}
\usepackage{generic}
\usepackage{cite}
\usepackage{amsmath,amssymb,amsfonts}
\usepackage{graphicx}
\usepackage{algorithm}
\usepackage{hyperref}
\usepackage{algpseudocode}
\usepackage{graphicx}
\usepackage{textcomp}
\usepackage{xcolor}
\usepackage{colortbl}
\usepackage{multirow}
\usepackage{booktabs}
\usepackage{url}
\hypersetup{hidelinks=true}
\usepackage{textcomp}
\definecolor{myred}{HTML}{FF0000}
\definecolor{myblue}{HTML}{0000FF}

\newcommand{\Rmnum}[1]{\expandafter\@slowromancap\romannumeral #1@}
\newcommand{\best}[1]{\textbf{\textcolor{myred}{#1}}}
\newcommand{\secondbest}[1]{\textbf{\textcolor{myblue}{#1}}}
\newcommand{\normval}[1]{#1} 
\def\BibTeX{{\rm B\kern-.05em{\sc i\kern-.025em b}\kern-.08em
    T\kern-.1667em\lower.7ex\hbox{E}\kern-.125emX}}
\begin{document}
\title{EVM-Fusion: An Explainable Vision Mamba Architecture with Neural Algorithmic Fusion}
\author{Zichuan Yang, Yongzhi Wang
\thanks{This research was funded by the National Undergraduate Innovation and Entrepreneurship Training Program under Grant No. 202410247022.}
\thanks{Zichuan Yang, School of Mathematical Sciences, Tongji University, Shanghai 200092 China, (e-mail: 2153747@tongji.edu.cn).}
\thanks{Yongzhi Wang, School of Mathematical Sciences, Tongji University, Shanghai 200092 China, (e-mail: 98713@tongji.edu.cn).}}

\maketitle

\begin{abstract}
Medical image classification is critical for clinical decision-making, yet demands for accuracy, interpretability, and generalizability remain challenging. This paper introduces EVM-Fusion, an Explainable Vision Mamba architecture featuring a novel Neural Algorithmic Fusion (NAF) mechanism for multi-organ medical image classification. EVM-Fusion leverages a multipath design, where DenseNet and U-Net based pathways, enhanced by Vision Mamba (Vim) modules, operate in parallel with a traditional feature pathway. These diverse features are dynamically integrated via a two-stage fusion process: cross-modal attention followed by the iterative NAF block, which learns an adaptive fusion algorithm. Intrinsic explainability is embedded through path-specific spatial attention, Vim $\Delta$ -value maps, traditional feature SE-attention, and cross-modal attention weights. Experiments on a diverse 9-class multi-organ medical image dataset demonstrate EVM-Fusion's strong classification performance and provide multi-faceted insights into its decision-making process, highlighting its potential for trustworthy AI in medical diagnostics.
\end{abstract}


\section{Introduction}
\label{sec:introduction}
\IEEEPARstart{T}{he} accurate analysis of medical images is fundamental to modern clinical practice \cite{bib34,bib2,bib4,bib5}. A standard and effective workflow for detailed diagnosis involves two stages. First, a model segments \cite{bib14} a specific region of interest (ROI). Then, a second model classifies that ROI. This segmentation-based approach is precise \cite{bib15,bib17,bib18}. However, it requires detailed pixel-level annotations, which are often expensive and time-consuming to create.

This paper explores a different but equally important clinical problem: automated image triage. A triage system needs to make a rapid, holistic assessment of an entire medical image. Its goal is to quickly categorize the image, such as identifying a potential abnormality or its likely type. This can help clinicians prioritize urgent cases and manage workflows more efficiently, especially in settings with a high volume of patients.

Building an effective end-to-end model for this triage task is challenging. The model must analyze the whole image without explicit guidance. This creates three specific problems. First, the model needs to understand both fine-grained local details and long-range global context. For example, it must see both the texture of a small lesion and its position relative to the entire organ. Standard Convolutional Neural Networks (CNNs) \cite{bib4,bib22,bib23} are good at local feature extraction but often struggle to capture global context efficiently. Second, a single image contains very different kinds of information. Deep learning models extract powerful hierarchical features. At the same time, traditional handcrafted features can provide robust and explicit information about texture or statistics \cite{bib3,bib6,bib1,bib10,bib11}. An effective triage model should use all these heterogeneous feature types. However, fusing them is a difficult problem. Simple methods like feature concatenation often fail because they cannot model the complex, non-linear relationships between these different feature streams. Third, many complex deep learning models operate like a "black box". This lack of transparency is a major barrier to trust and adoption in clinical settings. A reliable triage system must be interpretable, allowing clinicians to understand the basis of its decisions.

To solve these specific challenges, we designed and validated the Explainable Vision Mamba Architecture with Neural Algorithmic Fusion (EVM-Fusion). Our work is not about creating a single new component. Instead, our contribution is a synergistic framework that intelligently integrates multiple advanced components. Each component is chosen to address one of the problems mentioned above. We propose: \textbf{A Mamba-enhanced multi-path architecture to capture heterogeneous features.} The framework has three paths. A DenseNet path extracts robust semantic features. A U-Net path focuses on multi-scale details. A third path provides traditional texture features. We enhance the two deep learning paths with Vision Mamba (Vim) \cite{bib35} modules to specifically address the challenge of modeling long-range dependencies. \textbf{A learnable, two-stage fusion strategy to integrate these diverse features.} A Cross-Modal Attention stage first lets the different paths exchange contextual information. Then, a Neural Algorithmic Fusion (NAF) block learns a multi-step, iterative process to find the best way to combine the features. This design directly addresses the limitations of simpler, fixed fusion rules. \textbf{A holistically integrated explainability (XAI) system.} XAI \cite{bib37} mechanisms are built into every stage of the model. This provides a multi-faceted view of the model's reasoning process, which is essential for building a trustworthy AI tool for clinicians.

We tested our framework on a challenging 9-class dataset. Our experiments show that the complete EVM-Fusion system outperforms powerful state-of-the-art models from different architectural families. Our ablation studies further prove that every component in our synergistic design is necessary for achieving this top performance. The remainder of this paper is organized as follows: Section~\ref{sec:Related} reviews related work. Section~\ref{sec:model} details the EVM-Fusion architecture. Section~\ref{sec:experiment} describes the experimental setup. Section~\ref{sec:discussion} presents and discusses the results, and Section~\ref{sec:conclusion} concludes the paper.

\section{Related Work}
\label{sec:Related}

The field of medical image analysis has two main paradigms. The first is a two-stage, segmentation-based approach. Models like U-Net \cite{bib20} are often used to delineate specific ROIs \cite{bib26,bib27,bib28,bib29}, which are then classified. This method is the standard for precise clinical diagnosis. The second paradigm is holistic, end-to-end classification \cite{bib7,bib8,bib9,bib12,bib13}. Here, a model processes the entire image directly to make a prediction. Our work contributes to this second paradigm. We aim to develop a robust framework for rapid image triage, a task where holistic analysis can be more practical than detailed segmentation.

\subsection{Architectures for Visual Representation}
CNNs like ResNet \cite{bib24} and DenseNet \cite{bib19} have been foundational in medical imaging. They are very effective at learning local patterns from images. However, their fixed and local receptive fields are a known limitation for modeling global context. Vision Transformers (ViT) \cite{bib31} and Swin Transformer \cite{bib38} address this problem using a self-attention mechanism from natural language processing. More recently, State Space Models (SSMs) have emerged as another powerful tool for sequence modeling. Mamba \cite{bib46} is a key example. Vision-specific versions like VMamba \cite{bib45} and Vim adapt this technology for images. They can capture long-range dependencies with high efficiency. Our framework recognizes the distinct strengths of these architectures. We use CNNs for their proven ability to extract local features and enhance them with Mamba to add the crucial capability of global context modeling.

\subsection{Feature Fusion in Deep Learning}
Combining features from multiple sources is a common strategy to improve model performance. Simple approaches include feature concatenation or element-wise averaging. These methods are straightforward but may not be optimal \cite{bib25}. More advanced techniques use attention mechanisms to dynamically assign weights to different features \cite{bib50}. However, these methods typically use a single-step weighting scheme. This can be too rigid for fusing features that are highly diverse, such as combining deep hierarchical features with traditional statistical features. This specific challenge motivates our use of Neural Algorithmic Fusion (NAF) \cite{bib21}. NAF learns a more complex, multi-step fusion process, which is better suited for this difficult task.

\subsection{Explainable AI (XAI) in Medical Imaging}
The "black box" nature of many deep learning models hinders their clinical adoption, making XAI crucial \cite{bib39}. Common post-hoc XAI methods include LIME \cite{bib40} and SHAP \cite{bib41}. For CNNs, activation mapping techniques like CAM \cite{bib42} and its derivatives (Grad-CAM, Grad-CAM++) \cite{bib43} highlight image regions influencing predictions. However, building intrinsically interpretable models, where explainability is a core part of the architecture, is an increasingly important research direction \cite{bib44}. Our EVM-Fusion incorporates multiple intrinsic XAI components at different stages.

\begin{figure*}[ht]
    \centering
    \includegraphics[width=\textwidth, height=0.2\textheight, keepaspectratio]{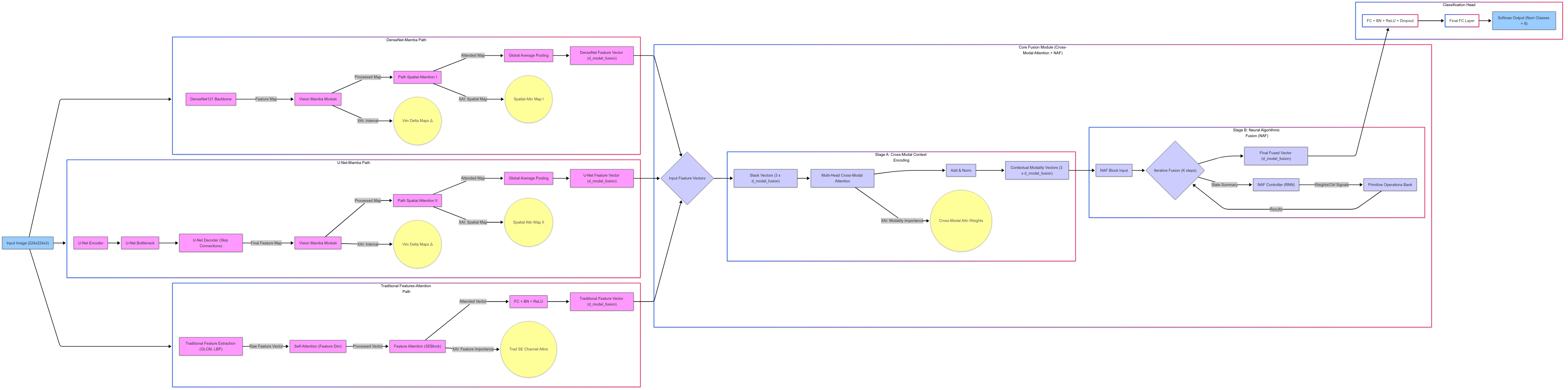}
    \caption{The proposed EVM-Fusion architecture. The model takes a single image as input and processes it through three parallel feature extraction and enhancement paths: (1) a DenseNet-Mamba path leveraging a DenseNet121 backbone followed by a Vision Mamba module and path-specific spatial attention; (2) a U-Net-Mamba path employing a full U-Net encoder-decoder structure, with a Vision Mamba module applied to the final decoder output, followed by spatial attention; and (3) a Traditional Features-Attention path that extracts GLCM and LBP features, subsequently refined by self-attention and Squeeze-and-Excitation (SE) attention. Feature vectors from these paths are then processed by a two-stage Core Fusion Module, which first uses Cross-Modal Attention for contextual encoding of inter-path relationships, followed by a novel Neural Algorithmic Fusion (NAF) block for iterative, adaptive fusion. The final fused vector is passed to a classification head. Intrinsic Explainable AI (XAI) outputs, including spatial attention maps, Vim $\Delta$-value maps, traditional feature importance scores, and cross-modal attention weights, are generated at various stages to provide model interpretability.}
    \label{Fig.1}
\end{figure*}

\section{Model Architecture}
\label{sec:model}

The EVM-Fusion architecture is shown in Fig~\ref{Fig.1}. It is a multi-path network. We designed it to address the challenges of holistic image triage. A key part of our design is the two-stage fusion strategy. This strategy uses a Neural Algorithmic Fusion (NAF) block to learn an iterative process for combining information. The architecture also has intrinsic XAI components to ensure transparency.

\subsection{Overall Framework}

Given an input medical image $\mathbf{I} \in \mathbb{R}^{H\times W\times C}$ , EVM-Fusion employs three parallel pathways to extract a rich set of complementary features. The motivation for this multi-path design is to capture different facets of the image data simultaneously: A DenseNet-Mamba path: Designed to capture robust, high-level semantic features by leveraging the deep feature reuse of DenseNet. A U-Net-Mamba path: Focused on extracting multi-scale spatial hierarchies and preserving fine-grained local details, a strength of the U-Net architecture's encoder-decoder structure. A Traditional Features-Attention path: Included to provide explicit, domain-relevant texture and statistical features that deep models might overlook.

Each path $p$ produces a feature vector $\mathbf{v}_p \in \mathbb{R}^{d_{\text{model\_fusion}}}$. These vectors are then passed to the Core Fusion Module, which first applies Cross-Modal Attention to model inter-path dependencies, followed by the NAF block to generate a final, unified feature representation $\mathbf{v}_{\text{fused}}\in \mathbb{R}^{d_{\text{model\_fusion}}}$. This vector is then used by a classification head to predict the image class.

\subsection{Feature Extraction and Enhancement Paths}

Each deep learning path incorporates a Vision Mamba module for advanced sequence-based feature refinement and a path-specific spatial attention mechanism.

\subsubsection{DenseNet-Mamba Path}

This path aims to extract robust semantic features. The input image $\mathbf{I}$ is first processed by a pre-trained DenseNet-121 backbone to obtain a feature map $\mathbf{F}_D=DenseNet(\mathbf{I})\in \mathbb{R}^{H_D\times W_D\times C_D}$. To model long-range dependencies, this feature map is processed by a Vision Mamba (Vim) module. It first transforms $\mathbf{F}_D$ into a sequence of patch tokens $\mathbf{X}_D\in\mathbb{R}^{N_p\times d_{mamba}}$. The core of the module consists of stacked Mamba (S6) blocks \cite{bib46} that process the sequence. In brief, for the $k$-th patch token $\mathbf{x}_k$, a Mamba block computes an output $\mathbf{y}_k$ via a selective state space model. The key state update is:
\begin{equation}
\mathbf{h}_k = \bar{\mathbf{A}}_k\mathbf{h}_{k-1} + \bar{\mathbf{B}}_k\mathbf{x}_k''
\end{equation}
Where $\mathbf{x}_k''$ is a projection of the input $\mathbf{x}_k$. The matrices $\bar{\mathbf{A}}_k$ and $\bar{\mathbf{B}}_k$ are parameterized by a data-dependent scalar $\mathbf{\Delta}_k$, which acts as a selective gate. This gate modulates the flow of information, allowing the model to either focus on the current input or maintain its previous state, effectively learning long-range dependencies. The $\mathbf{\Delta}_k$ values are stored for XAI analysis. The Vim module's output is reshaped back into a feature map $\mathbf{F}_{D,map}' \in\mathbb{R}^{H_D'\times W_D'\times d_{\text{model\_fusion}}}$.

To recalibrate the feature map spatially, a spatial attention module is applied:
\begin{equation}
\begin{split}
\mathbf{M}_{\text{spat}}(\mathbf{F'}) = \sigma(\mathbf{W}_{\text{conv}}[\text{AvgPool}_{\text{ch}}(\mathbf{F'});\\ \text{MaxPool}_{\text{ch}}(\mathbf{F'})])
\end{split}
\end{equation}
where channel-wise average and max pooling aggregate feature information, and $\sigma$ is the sigmoid function. The final path vector is $\mathbf{v}_D=GAP(\mathbf{F}'\odot\mathbf{M}_{\text{spat}}(\mathbf{F}'))$, where GAP is Global Average Pooling. The attention map $\mathbf{M}_{\text{spat}}$ is saved for XAI.

\subsubsection{U-Net-Mamba Path}

This path is designed to capture multi-scale spatial information and fine-grained details, which are often critical for medical image analysis. It employs a full U-Net architecture as the initial feature extractor. The output from the U-Net's final decoder layer, a feature map $\mathbf{F}_U$, is then processed in a manner identical to the DenseNet path: it is passed through its own dedicated Vision Mamba and Spatial Attention modules to produce the feature vector $\mathbf{v}_D\in\mathbb{R}^{d_{\text{model\_fusion}}}$.

\subsubsection{Traditional Features-Attention Path}

This path explicitly incorporates domain-relevant handcrafted features. For a grayscale version of input $\mathbf{I}$, Gray-Level Co-occurrence Matrix (GLCM) and Local Binary Pattern (LBP) features are extracted into a raw feature vector $\mathbf{f}_{raw} \in \mathbb{R}^{d_{raw}}$.

As these features lack a natural sequential order, a Multi-Head Attention (MHA) layer \cite{bib47} is used instead of Mamba. Here, MHA learns inter-feature relationships, allowing features to be re-weighted based on the context of other features in the vector.
\begin{equation}
\mathbf{f}'_{\text{trad}} = \text{MHA}(\mathbf{f}_{\text{raw}},\mathbf{f}_{\text{raw}},\mathbf{f}_{\text{raw}})
\end{equation}
This is followed by a Squeeze-and-Excitation (SE) block \cite{bib48} to perform feature-wise recalibration:
\begin{equation}
\mathbf{s} = \sigma(\mathbf{W}_2\text{ReLU}(\mathbf{W}_1\mathbf{f}'_{\text{trad}}))
\end{equation}
Where $\mathbf{W}_1\in\mathbb{R}^{(d_{raw}/r)\times d_{raw}}$ and $\mathbf{W}_2\in\mathbb{R}^{d_{raw}\times (d_{raw}/r)}$ are linear layers, $r$ is the reduction ratio. The output $\mathbf{f}_{trad}''=\mathbf{s}\odot \mathbf{f}_{trad}'$
 , and the scores s are saved for XAI. Finally, a fully connected layer produces $\mathbf{v}_T\in \mathbb{R}^{d_{\text{model\_fusion}}}$.

\subsection{Core Fusion Module}

This module intelligently integrates the feature vectors $\mathbf{v}_D, \mathbf{v}_U, \mathbf{v}_T$.

\subsubsection{Stage A: Cross-Modal Context Encoding}\label{subsubsec4}

To enable the different feature streams to inform one another before fusion, I first stack them into a matrix $\mathbf{v}_{in}=[\mathbf{v}_D;\mathbf{v}_U;\mathbf{v}_T]\in\mathbb{R}^{N_{paths}\times d_{model\_fusion}}$, where $N_{paths}=3$. Then $V_{in}$ is processed by an MHA layer acting as a cross-modal attention mechanism.
\begin{equation}
\mathbf{V}_{\text{attended}}, \mathbf{W}_{\text{CMA}} = \text{MHA}(\mathbf{V}_{\text{in}}, \mathbf{V}_{\text{in}}, \mathbf{V}_{\text{in}})
\end{equation}
The attention weights $\mathbf{W}_{\text{CMA}}\in\mathbb{R}^{N_{paths}\times N_{paths}}$ are saved for XAI. A residual connection followed by Layer Normalization is applied to produce the contextualized feature matrix:
\begin{equation}
\mathbf{V}_{\text{contextual}} = \text{LayerNorm}(\mathbf{V}_{\text{in}} + \mathbf{V}_{\text{attended}})
\end{equation}

\begin{table*}[htbp]
\centering
\caption{Dataset Summary}
\label{tab:1}
\resizebox{\textwidth}{!}{%
\begin{tabular}{|c|cccc|cc|ccc|c|}
\hline
\textbf{} & \multicolumn{4}{c|}{\textbf{Brain Tumor}} & \multicolumn{2}{c|}{\textbf{Breast Cancer}} & \multicolumn{3}{c|}{\textbf{Chest X-Ray}} & \textbf{Sum} \\
\textbf{} & glioma & healthy & meningioma & pituitary & 0 & 1 & normal & bacteria & virus &  \\
\hline
train & 1134 & 1400 & 1151 & 1229 & 1569 & 803 & 1145 & 2159 & 1148 & 11638 \\
validation & 243 & 300 & 246 & 263 & 448 & 227 & 204 & 379 & 197 & 2507 \\
test & 244 & 300 & 248 & 265 & 208 & 128 & 234 & 242 & 148 & 2017 \\
\hline
sum & 1621 & 2000 & 1645 & 1757 & 2225 & 1158 & 1583 & 2780 & 1493 & 16162 \\
\hline
\end{tabular}%
}
\end{table*}

\subsubsection{Stage B: Neural Algorithmic Fusion (NAF) Block}

To move beyond a single-step fusion, the NAF block is designed to learn an iterative refinement algorithm. Its strength lies in a controller-driven process that dynamically mixes the outputs of several simple "primitive" operations.

The process is initialized for $K_{\text{NAF}}$ iterations. At each step $k$, a controller (a GRU \cite{bib49}) receives the current fused state $\mathbf{S}_{fused}^{(k-1)}$ and generates mixing weights $\alpha^{(k)}$ for a bank of $N_{primitives}$ primitive operations $\{\mathcal{P}_j\}$. Each primitive (a simple MLP) processes the static, comprehensive feature set $\mathbf{V}_{\text{contextual}}$. The outputs of the primitives are then mixed according to $\alpha^{(k)}$ to produce an update, which is added to the current state via a residual connection.

\begin{algorithm}[h!]
\caption{Neural Algorithmic Fusion (NAF) Block}
\label{alg:naf}
\begin{algorithmic}[1]
    \Require 
        A set of contextualized feature vectors $\{\mathbf{v}_i\}_{i=1}^{N_{\text{paths}}}$, where $\mathbf{v}_i \in \mathbb{R}^{d_{\text{model\_fusion}}}$.
    \Ensure 
        The final fused feature vector $\mathbf{v}_{\text{NAF}} \in \mathbb{R}^{d_{\text{model\_fusion}}}$.
    
    \State $\mathbf{v}_{\text{cat}} \leftarrow \text{Concatenate}(\mathbf{v}_1, \dots, \mathbf{v}_{N_{\text{paths}}})$ \Comment{Concatenate features from all paths}
    \State $\mathbf{s}_{\text{fused}} \leftarrow \text{Linear}(\mathbf{v}_{\text{cat}})$ \Comment{Project to the internal state dimension $d_{\text{state}}$}
    \State $\mathbf{h}_{\text{ctrl}} \leftarrow \mathbf{0} \in \mathbb{R}^{d_{\text{ctrl}}}$ \Comment{Initialize controller (GRU) hidden state}
    
    \For{$j = 1 \to N_{\text{primitives}}$} \Comment{Pre-compute primitives from static input}
        \State $\mathbf{p}_j \leftarrow \text{Primitive}_j(\mathbf{v}_{\text{cat}})$ \Comment{Each $\text{Primitive}_j$ is a small MLP; $\mathbf{p}_j \in \mathbb{R}^{d_{\text{state}}}$}
    \EndFor
    
    \For{$k = 1 \to K_{\text{NAF}}$} \Comment{Iteratively refine the fused state}
        \State $\mathbf{h}_{\text{ctrl}} \leftarrow \text{GRU}(\mathbf{s}_{\text{fused}}, \mathbf{h}_{\text{ctrl}})$ \Comment{Controller observes current state and updates}
        \State $\mathbf{\alpha} \leftarrow \text{Softmax}(\text{Linear}(\mathbf{h}_{\text{ctrl}}))$ \Comment{Generate mixing weights $\mathbf{\alpha} \in \mathbb{R}^{N_{\text{primitives}}}$}
        \State $\mathbf{s}_{\text{mix}} \leftarrow \sum_{j=1}^{N_{\text{primitives}}} \alpha_j \mathbf{p}_j$ \Comment{Calculate weighted sum of primitives; $\mathbf{s}_{\text{mix}} \in \mathbb{R}^{d_{\text{state}}}$}
        \State $\mathbf{s}_{\text{fused}} \leftarrow \text{LayerNorm}(\mathbf{s}_{\text{fused}} + \mathbf{s}_{\text{mix}})$ \Comment{Update state with residual connection}
    \EndFor
    
    \State $\mathbf{v}_{\text{NAF}} \leftarrow \text{Linear}(\mathbf{s}_{\text{fused}})$ \Comment{Project final state to the output dimension $d_{\text{model\_fusion}}$}
\end{algorithmic}
\end{algorithm}

\begin{table*}[!htbp]
\centering
\caption{Ablation results: Detailed comparison of performance metrics and per-category scores after removing some modules from the model. Best values in each column are in \textbf{\textcolor{myred}{red}}, second best in \textbf{\textcolor{myblue}{blue}}.}
\label{tab:2}
\resizebox{\textwidth}{!}{%
\begin{tabular}{@{}lcccccccccccccccccccccccccccccccc@{}}
\toprule
\multirow{2}{*}{\textbf{Model}} & \multicolumn{4}{c}{\textbf{Overall}} & \multicolumn{3}{c}{\textbf{0}} & \multicolumn{3}{c}{\textbf{1}} & \multicolumn{3}{c}{\textbf{Bacteria}} & \multicolumn{3}{c}{\textbf{Glioma}} & \multicolumn{3}{c}{\textbf{Healthy}} & \multicolumn{3}{c}{\textbf{Meningioma}} & \multicolumn{3}{c}{\textbf{Normal}} & \multicolumn{3}{c}{\textbf{Pituitary}} & \multicolumn{3}{c}{\textbf{Virus}} \\
\cmidrule(lr){2-5} \cmidrule(lr){6-8} \cmidrule(lr){9-11} \cmidrule(lr){12-14} \cmidrule(lr){15-17} \cmidrule(lr){18-20} \cmidrule(lr){21-23} \cmidrule(lr){24-26} \cmidrule(lr){27-29} \cmidrule(lr){30-32}
 & Acc & P & R & F1 & P & R & F1 & P & R & F1 & P & R & F1 & P & R & F1 & P & R & F1 & P & R & F1 & P & R & F1 & P & R & F1 & P & R & F1 \\
\midrule
(Proposed Method) EVM-Fusion (DUHF) & \best{0.9479} & \best{0.9557} & \best{0.9479} & \best{0.9484} & \best{0.9351} & \normval{0.6923} & \best{0.7956} & \secondbest{0.6484} & \best{0.9219} & \best{0.7613} & \best{0.9747} & \normval{0.9545} & \best{0.9645} & \best{1.0000} & \best{1.0000} & \best{1.0000} & \secondbest{0.9967} & \best{1.0000} & \best{0.9983} & \best{1.0000} & \best{0.9960} & \best{0.9980} & \normval{0.9576} & \best{0.9658} & \best{0.9617} & \best{1.0000} & \best{1.0000} & \best{1.0000} & \best{0.9073} & \best{0.9257} & \best{0.9164} \\
(Path Ablation) EVM-Fusion (DHF) & \normval{0.8820} & \normval{0.8907} & \normval{0.8820} & \normval{0.8809} & \normval{0.7193} & \normval{0.7885} & \normval{0.7523} & \normval{0.5926} & \normval{0.5000} & \normval{0.5424} & \normval{0.8269} & \secondbest{0.9669} & \normval{0.8914} & \secondbest{0.9958} & \normval{0.9754} & \normval{0.9855} & \normval{0.9934} & \best{1.0000} & \secondbest{0.9967} & \normval{0.9761} & \secondbest{0.9879} & \normval{0.9820} & \secondbest{0.9936} & \normval{0.6667} & \normval{0.7980} & \normval{0.9887} & \normval{0.9887} & \normval{0.9887} & \normval{0.6304} & \normval{0.7838} & \normval{0.6988} \\
(Path Ablation) EVM-Fusion (DU) & \normval{0.8934} & \normval{0.9028} & \normval{0.8934} & \normval{0.8945} & \secondbest{0.7556} & \secondbest{0.8173} & \secondbest{0.7852} & \best{0.6607} & \normval{0.5781} & \secondbest{0.6167} & \normval{0.8776} & \normval{0.8884} & \normval{0.8830} & \normval{0.9877} & \normval{0.9877} & \normval{0.9877} & \secondbest{0.9967} & \normval{0.9933} & \normval{0.9950} & \normval{0.9879} & \normval{0.9839} & \secondbest{0.9859} & \normval{0.9773} & \normval{0.7350} & \normval{0.8390} & \normval{0.9887} & \normval{0.9925} & \secondbest{0.9906} & \normval{0.6158} & \normval{0.8446} & \normval{0.7123} \\
(Path Ablation) EVM-Fusion (UHF) & \normval{0.7685} & \normval{0.8291} & \normval{0.7685} & \normval{0.7606} & \normval{0.3701} & \normval{0.7740} & \normval{0.5008} & \normval{0.4471} & \normval{0.2969} & \normval{0.3568} & \normval{0.9069} & \normval{0.9256} & \normval{0.9162} & \normval{0.9138} & \normval{0.2172} & \normval{0.3510} & \normval{0.9700} & \normval{0.9700} & \normval{0.9700} & \normval{0.9194} & \normval{0.9194} & \normval{0.9194} & \normval{0.9706} & \normval{0.7051} & \normval{0.8168} & \normval{0.9630} & \normval{0.9811} & \normval{0.9720} & \normval{0.6373} & \secondbest{0.8784} & \normval{0.7386} \\
(Path Ablation) DenseNet\cite{bib19} & \secondbest{0.8969} & \normval{0.8986} & \secondbest{0.8969} & \secondbest{0.8954} & \normval{0.7021} & \normval{0.7933} & \normval{0.7449} & \normval{0.5743} & \normval{0.4531} & \normval{0.5066} & \normval{0.8613} & \best{0.9752} & \normval{0.9147} & \normval{0.9835} & \normval{0.9754} & \normval{0.9794} & \best{1.0000} & \secondbest{0.9967} & \best{0.9983} & \normval{0.9679} & \normval{0.9718} & \normval{0.9698} & \best{0.9948} & \normval{0.8120} & \secondbest{0.8941} & \normval{0.9813} & \normval{0.9887} & \normval{0.9850} & \normval{0.7547} & \normval{0.8108} & \secondbest{0.7818} \\
(Path Ablation) U-Net\cite{bib20} & \normval{0.8374} & \normval{0.8471} & \normval{0.8374} & \normval{0.8378} & \normval{0.6937} & \normval{0.5337} & \normval{0.6033} & \normval{0.4489} & \normval{0.6172} & \normval{0.5197} & \normval{0.8206} & \normval{0.8884} & \normval{0.8532} & \normval{0.9424} & \normval{0.9385} & \normval{0.9405} & \normval{0.9637} & \normval{0.9733} & \normval{0.9685} & \normval{0.9469} & \normval{0.8629} & \normval{0.9030} & \normval{0.8956} & \normval{0.6966} & \normval{0.7837} & \normval{0.9059} & \normval{0.9811} & \normval{0.9420} & \normval{0.7079} & \normval{0.8514} & \normval{0.7730} \\
(Path Ablation) U-DenseNet\cite{bib26} & \normval{0.8537} & \normval{0.8526} & \normval{0.8537} & \normval{0.8495} & \normval{0.6484} & \normval{0.7981} & \normval{0.7155} & \normval{0.4750} & \normval{0.2969} & \normval{0.3654} & \normval{0.8297} & \normval{0.9463} & \normval{0.8842} & \normval{0.9732} & \normval{0.8934} & \normval{0.9316} & \normval{0.9767} & \normval{0.9767} & \normval{0.9767} & \normval{0.8716} & \normval{0.9032} & \normval{0.8871} & \normval{0.9439} & \normval{0.7906} & \normval{0.8605} & \normval{0.9348} & \normval{0.9736} & \normval{0.9538} & \normval{0.7303} & \normval{0.7500} & \normval{0.7400} \\
(Fusion Ablation) Simple-Mean & \normval{0.8880} & \normval{0.8930} & \normval{0.8880} & \normval{0.8869} & \normval{0.6933} & \normval{0.7933} & \normval{0.7399} & \normval{0.5657} & \normval{0.4375} & \normval{0.4934} & \normval{0.8902} & \normval{0.9380} & \normval{0.9135} & \normval{0.9758} & \secondbest{0.9918} & \normval{0.9837} & \normval{0.9966} & \normval{0.9900} & \normval{0.9933} & \normval{0.9757} & \normval{0.9718} & \normval{0.9737} & \normval{0.9721} & \normval{0.7436} & \normval{0.8426} & \secondbest{0.9924} & \normval{0.9849} & \normval{0.9886} & \normval{0.6737} & \normval{0.8649} & \normval{0.7574} \\
(Fusion Ablation) Simple-Concat & \normval{0.8721} & \normval{0.8662} & \normval{0.8721} & \normval{0.8531} & \normval{0.6211} & \best{0.9615} & \normval{0.7547} & \normval{0.4667} & \normval{0.0547} & \normval{0.0979} & \normval{0.8014} & \secondbest{0.9669} & \normval{0.8764} & \normval{0.9785} & \normval{0.9344} & \normval{0.9560} & \normval{0.9900} & \normval{0.9867} & \normval{0.9883} & \normval{0.9469} & \normval{0.9355} & \normval{0.9412} & \normval{0.9412} & \secondbest{0.8205} & \normval{0.8767} & \normval{0.9429} & \secondbest{0.9962} & \normval{0.9688} & \secondbest{0.8346} & \normval{0.7162} & \normval{0.7709} \\
(Fusion Ablation) NAF-Only & \normval{0.8681} & \normval{0.8799} & \normval{0.8681} & \normval{0.8682} & \normval{0.5860} & \normval{0.8029} & \normval{0.6775} & \normval{0.5233} & \normval{0.3516} & \normval{0.4206} & \normval{0.8731} & \normval{0.9380} & \normval{0.9044} & \normval{0.9916} & \normval{0.9713} & \normval{0.9814} & \best{1.0000} & \normval{0.8900} & \normval{0.9418} & \normval{0.9683} & \normval{0.9839} & \normval{0.9760} & \normval{0.9778} & \normval{0.7521} & \normval{0.8502} & \normval{0.9887} & \normval{0.9925} & \secondbest{0.9906} & \normval{0.6868} & \normval{0.8446} & \normval{0.7576} \\
(Fusion Ablation) CMA-Only & \normval{0.8934} & \secondbest{0.9041} & \normval{0.8934} & \secondbest{0.8954} & \normval{0.7500} & \normval{0.6346} & \normval{0.6875} & \normval{0.5276} & \secondbest{0.6719} & \normval{0.5911} & \secondbest{0.9036} & \normval{0.9298} & \secondbest{0.9165} & \normval{0.9879} & \best{1.0000} & \secondbest{0.9939} & \normval{0.9966} & \normval{0.9900} & \normval{0.9933} & \secondbest{0.9918} & \normval{0.9798} & \normval{0.9858} & \normval{0.9892} & \normval{0.7821} & \normval{0.8735} & \normval{0.9887} & \normval{0.9887} & \normval{0.9887} & \normval{0.6878} & \secondbest{0.8784} & \normval{0.7715} \\
\bottomrule
\end{tabular}}
\end{table*}

The controller's state update at step k follows standard GRU equations:
\begin{equation}
\mathbf{h}_{\text{ctrl}}^{(k)} = \text{GRU}(\mathbf{S}_{\text{fused}}^{(k-1)}, \mathbf{h}_{\text{ctrl}}^{(k-1)})
\end{equation}
where$\mathbf{h}_{\text{ctrl}}^{(k)}$ is the GRU's hidden state. This state is used to generate mixing weights:
\begin{equation}
\mathbf{\alpha}^{(k)} = \text{softmax}(\mathbf{W}_{\text{mix}}\mathbf{h}_{\text{ctrl}}^{(k)} + \mathbf{b}_{\text{mix}})
\end{equation}

A bank of primitives processes $\mathbf{V}_{\text{contextual}}$ (here flattened and concatenated), and their outputs are mixed:
\begin{equation}
\mathbf{S}_{\text{mix}}^{(k)} = \sum_{j=1}^{N_{\text{primitives}}} \alpha_j^{(k)} \mathcal{P}_j(\mathbf{V}_{\text{contextual}})
\end{equation}
The fused state is then updated with a residual connection:
\begin{equation}
\mathbf{S}_{\text{fused}}^{(k)} = \text{LayerNorm}(\mathbf{S}_{\text{fused}}^{(k-1)} + \mathbf{S}_{\text{mix}}^{(k)})
\end{equation}

After $K_{NAF}$ iterations, the final state $\mathbf{S}_{\text{fused}}^{(K_{NAF})}$ is processed by a linear layer to produce the final fused vector $\mathbf{V}_{\text{NAF}}$.

\begin{table*}[!htbp]
\centering
\caption{Detailed Comparison of Model Performance Metrics with Per-Class Scores and Parameters. Best values are in \textbf{\textcolor{myred}{red}}, second best in \textbf{\textcolor{myblue}{blue}}.}
\label{tab:3}
\resizebox{\textwidth}{!}{%
\begin{tabular}{@{}l|c|cccccccccccccccccccccccccccccccc@{}}
\toprule
\multirow{2}{*}{\textbf{Model}} & \multirow{2}{*}{\textbf{Params (M)}} & \multicolumn{4}{c}{\textbf{Overall}} & \multicolumn{3}{c}{\textbf{0}} & \multicolumn{3}{c}{\textbf{1}} & \multicolumn{3}{c}{\textbf{Bacteria}} & \multicolumn{3}{c}{\textbf{Glioma}} & \multicolumn{3}{c}{\textbf{Healthy}} & \multicolumn{3}{c}{\textbf{Meningioma}} & \multicolumn{3}{c}{\textbf{Normal}} & \multicolumn{3}{c}{\textbf{Pituitary}} & \multicolumn{3}{c}{\textbf{Virus}} \\
\cmidrule(lr){3-6} \cmidrule(lr){7-9} \cmidrule(lr){10-12} \cmidrule(lr){13-15} \cmidrule(lr){16-18} \cmidrule(lr){19-21} \cmidrule(lr){22-24} \cmidrule(lr){25-27} \cmidrule(lr){28-30} \cmidrule(lr){31-33}
& & Acc & P & R & F1 & P & R & F1 & P & R & F1 & P & R & F1 & P & R & F1 & P & R & F1 & P & R & F1 & P & R & F1 & P & R & F1 & P & R & F1 \\
\midrule
(Proposed Method) EVM-Fusion & \normval{134.28} & \best{0.9479} & \best{0.9557} & \best{0.9479} & \best{0.9484} & \best{0.9351} & \normval{0.6923} & \secondbest{0.7956} & \normval{0.6484} & \best{0.9219} & \best{0.7613} & \best{0.9747} & \normval{0.9545} & \best{0.9645} & \best{1.0000} & \best{1.0000} & \best{1.0000} & \secondbest{0.9967} & \best{1.0000} & \best{0.9983} & \best{1.0000} & \best{0.9960} & \best{0.9980} & \best{0.9576} & \best{0.9658} & \best{0.9617} & \best{1.0000} & \best{1.0000} & \best{1.0000} & \best{0.9073} & \best{0.9257} & \best{0.9164} \\
Swim Transformer\cite{bib38} & \normval{195.01} & \normval{0.8136} & \normval{0.8215} & \normval{0.8136} & \normval{0.8146} & \normval{0.6859} & \normval{0.6298} & \normval{0.6566} & \normval{0.4690} & \normval{0.5312} & \normval{0.4982} & \normval{0.8359} & \normval{0.9050} & \normval{0.8690} & \normval{0.8996} & \normval{0.8811} & \normval{0.8903} & \normval{0.9801} & \normval{0.9867} & \normval{0.9834} & \normval{0.8142} & \normval{0.8306} & \normval{0.8224} & \normval{0.8864} & \normval{0.6667} & \normval{0.7610} & \normval{0.9046} & \normval{0.8943} & \normval{0.8994} & \normval{0.6043} & \normval{0.7635} & \normval{0.6746} \\
Internimage\cite{bib53} & \normval{221.43} & \normval{0.8934} & \normval{0.9028} & \normval{0.8934} & \normval{0.8945} & \normval{0.7556} & \normval{0.8173} & \normval{0.7852} & \normval{0.6607} & \normval{0.5781} & \normval{0.6167} & \normval{0.8776} & \normval{0.8884} & \normval{0.8830} & \normval{0.9877} & \normval{0.9877} & \normval{0.9877} & \secondbest{0.9967} & \secondbest{0.9933} & \secondbest{0.9950} & \normval{0.9879} & \normval{0.9839} & \normval{0.9859} & \normval{0.9773} & \normval{0.7350} & \normval{0.8390} & \normval{0.9887} & \normval{0.9925} & \normval{0.9906} & \normval{0.6158} & \normval{0.8446} & \normval{0.7123} \\
DeiT III\cite{bib51} & \normval{303.88} & \secondbest{0.9068} & \secondbest{0.9105} & \secondbest{0.9068} & \secondbest{0.9054} & \normval{0.7593} & \secondbest{0.8798} & \best{0.8151} & \best{0.7368} & \normval{0.5469} & \secondbest{0.6278} & \normval{0.8298} & \secondbest{0.9669} & \normval{0.8931} & \secondbest{0.9959} & \secondbest{0.9918} & \secondbest{0.9938} & \best{1.0000} & \normval{0.9867} & \normval{0.9933} & \secondbest{0.9880} & \secondbest{0.9919} & \secondbest{0.9899} & \normval{0.9840} & \secondbest{0.7906} & \secondbest{0.8768} & \secondbest{0.9888} & \best{1.0000} & \secondbest{0.9944} & \normval{0.6968} & \normval{0.7297} & \normval{0.7129} \\
MambaVision\cite{bib52} & \normval{226.41} & \normval{0.9033} & \normval{0.9073} & \normval{0.9033} & \normval{0.9003} & \normval{0.7160} & \best{0.8846} & \normval{0.7914} & \secondbest{0.6962} & \normval{0.4297} & \normval{0.5314} & \normval{0.8745} & \best{0.9793} & \secondbest{0.9240} & \normval{0.9877} & \normval{0.9877} & \normval{0.9877} & \best{1.0000} & \normval{0.9900} & \secondbest{0.9950} & \normval{0.9837} & \normval{0.9758} & \normval{0.9798} & \normval{0.9836} & \normval{0.7692} & \normval{0.8633} & \normval{0.9778} & \secondbest{0.9962} & \normval{0.9869} & \secondbest{0.7176} & \normval{0.8243} & \secondbest{0.7673} \\
ConvNeXt\cite{bib32} & \normval{196.24} & \normval{0.8954} & \normval{0.9067} & \normval{0.8954} & \normval{0.8976} & \secondbest{0.7598} & \normval{0.7452} & \normval{0.7524} & \normval{0.5985} & \secondbest{0.6172} & \normval{0.6077} & \secondbest{0.9348} & \normval{0.8884} & \normval{0.9110} & \normval{0.9877} & \normval{0.9877} & \normval{0.9877} & \best{1.0000} & \normval{0.9900} & \secondbest{0.9950} & \normval{0.9759} & \normval{0.9798} & \normval{0.9779} & \secondbest{0.9890} & \normval{0.7692} & \normval{0.8654} & \normval{0.9813} & \normval{0.9887} & \normval{0.9850} & \normval{0.6321} & \secondbest{0.9054} & \normval{0.7444} \\
\bottomrule
\end{tabular}
}
\end{table*}

\section{Experiments}
\label{sec:experiment}
We tested our model on a composite 9-class dataset. We created this dataset from three public sources: a brain tumor MRI dataset from Kaggle \cite{bib30}, a breast cancer X-ray dataset from Kaggle \cite{bib36}, and a chest X-ray pneumonia dataset from Mendeley Data \cite{bib16} proposed by Daniel S. Kermany et al. \cite{bib33}. This diverse dataset tests the model's ability to work with different modalities and organs. This is an important capability for a general-purpose triage system. Table~\ref{tab:1} shows the data distribution.

For a fair and rigorous comparison, all models were trained and evaluated under identical conditions. The baseline models were fine-tuned from their official ImageNet pretrained weights. The CNN backbones in our own model (DenseNet-121) were also initialized with ImageNet weights. This ensures that we are comparing the architectural advantages, not the benefits of pretraining.

\subsection{Comparison with State-of-the-Art Baselines}
A key part of our validation is comparing EVM-Fusion to strong, modern models. We chose a diverse set of baselines. Each one represents a different major architectural philosophy in computer vision. Table~\ref{tab:3} shows the results of this comparison, including overall and per-class metrics.

subsection{Ablation Studies: Validating the Synergistic Framework}
Our core hypothesis is that the strength of EVM-Fusion comes from the \textbf{synergy} between its components. To prove this, we conducted detailed ablation studies. The results are in Table~\ref{tab:2}. These studies were designed to answer two critical questions:

\textbf{Is each feature extraction path necessary?}: We tested the full model against versions where one or more of the paths were removed (e.g., EVM-Fusion (DHF) which removes the U-Net path).

\textbf{Is our two-stage fusion mechanism effective?}: We compared our full fusion module to simpler alternatives. These include Simple-Concat, Simple-Mean, CMA-Only, and NAF-Only.

\section{Discussion}
\label{sec:discussion}
The results in Table~\ref{tab:3} show that EVM-Fusion achieves state-of-the-art performance with an accuracy of 94.79\%. Our model consistently outperforms all selected baseline models. These baselines represent the best of different architectural families. This result suggests that for the complex task of holistic image triage, a single design philosophy is not enough. The success of EVM-Fusion comes from its ability to effectively integrate these different approaches into a single, synergistic framework.

The ablation studies in Table~\ref{tab:2} provide direct evidence for this synergy hypothesis. First, removing any of the three feature paths leads to a significant drop in performance. For example, removing the U-Net path (EVM-Fusion (DHF)) causes the accuracy to fall from 94.79\% to 88.20\%. This shows that the multi-scale features from the U-Net path provide unique and essential information.

Second, the fusion mechanism ablation is also very telling. The Simple-Concat baseline performs poorly (87.21\% accuracy). This confirms that just combining the features is not an effective strategy. Using CMA-Only or NAF-Only improves the result, but the full two-stage system (EVM-Fusion (DUHF)) is clearly the best. This validates our central claim. The best way to fuse these heterogeneous features is to first use CMA for contextualization, then use the powerful NAF block to learn a complex fusion algorithm.

\begin{figure*}
    \centering
    \includegraphics[width=\textwidth, height=0.2\textheight, keepaspectratio]{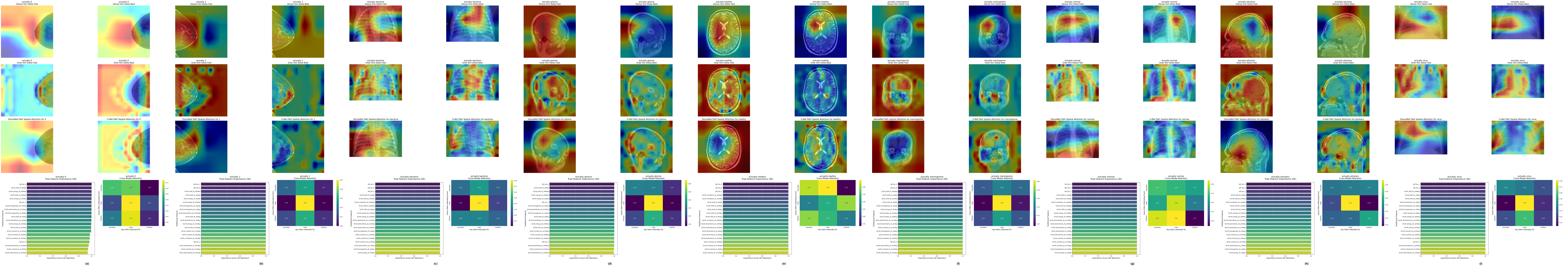}
    \caption{Explainable AI (XAI) Analysis of Representative Correctly Classified Test Samples by EVM-DUHF Across All Classes. This figure showcases the intrinsic explainability outputs for nine correctly classified samples, one from each of the distinct medical image classes in the test set. Noted from (a) to (i). Similar to Fig~\ref{Fig.2}, each sample is depicted through eight XAI visualizations. The analysis of these diverse, correctly classified instances aims to demonstrate the model's adaptive behavior, the varying contributions of its different pathways, and its ability to focus on salient, class-relevant features across multiple organ types and imaging modalities.} 
    \label{Fig.3}
\end{figure*}

Our XAI visualizations in Fig~\ref{Fig.3} provide a qualitative look at how this synergy works. \textbf{Vision Mamba's Adaptive Processing ($\Delta$-value Maps):} The $\Delta$-value maps from the Vision Mamba modules reveal dynamic, input-dependent information gating. Higher $\Delta$ values (warmer colors) signify a greater emphasis on current patch information, indicating regions of novelty or high salience. Conversely, lower $\Delta$ values (cooler colors) suggest reliance on previously propagated contextual states, often where information is redundant or predictable. For instance, in a correctly classified "normal" chest X-ray (Fig~\ref{Fig.3}g), the DenseNet path's Vim $\Delta$ maps show adaptive focus: the forward scan emphasizes lung boundaries upon encountering them, while the backward scan might highlight different boundary segments. The U-Net path's Vim modules on the same sample might show more detailed, localized $\Delta$ peaks within the lung fields. Comparing forward and backward $\Delta$ maps (e.g., Fig~\ref{Fig.3}b, a malignant tumor) often reveals differing "novelty" assessments based on the directional context of Mamba's selective scan, showcasing its capacity for nuanced contextual understanding. \textbf{Path-Specific Spatial Attention:} The spatial attention maps for the DenseNet and U-Net Mamba paths highlight prioritized image regions before global feature aggregation. In a correctly identified malignant tumor (Fig~\ref{Fig.3}b), these maps focus on the breast tissue, with DenseNet potentially emphasizing broader areas while U-Net targets more specific regions like the nipple, possibly indicating sensitivity to microcalcifications or lesion-dense areas. Interestingly, Fig~\ref{Fig.3}b also reveals a potential U-Net flaw, with some attention on blank image edges, underscoring the value of multi-path fusion to mitigate individual path imperfections. \textbf{Traditional Feature Importance (SE-Attention):} The Squeeze-and-Excitation (SE) attention weights within the traditional feature path quantify the contribution of specific GLCM and LBP features. For a correctly classified "normal" lung (Fig~\ref{Fig.3}g), features indicative of textural uniformity (LBP\_bin\_5 and LBP\_bin\_6) are often highly weighted. This demonstrates the model's ability to leverage well-understood handcrafted features that complement deep learned representations, capturing intrinsic textural properties relevant to the diagnostic class. \textbf{Cross-Modal Attention:} A cornerstone of EVM-Fusion's adaptability is the dynamic nature of its cross-modal attention weights, which vary significantly across different input images. For the benign tumors (Fig~\ref{Fig.3}a), all paths show increased attention towards the U-Net path, suggesting a reliance on its detailed local information for tumor characterization. Conversely, for a 'healthy' brain MRI (Fig~\ref{Fig.3}e), the attention shifts, with U-Net drawing more from the traditional path if specific textures are key. This input-dependent, flexible weighting allows EVM-Fusion to adaptively modulate the influence of each feature pathway prior to the final Neural Algorithmic Fusion (NAF) stage. Notably, across many samples in Fig~\ref{Fig.3}, most paths often receive substantial attention, indicating their collective importance and complementary roles. This inherent dynamism and balanced contribution underpin the superior performance of the full 3-path EVM-Fusion (EVM-DUHF) over ablated versions lacking one or more paths, as it allows the model to maximize information gain from diverse feature types and reduce over-reliance on any single potentially fallible source.

\begin{figure*}
    \centering
    \includegraphics[width=\textwidth, height=0.2\textheight, keepaspectratio]{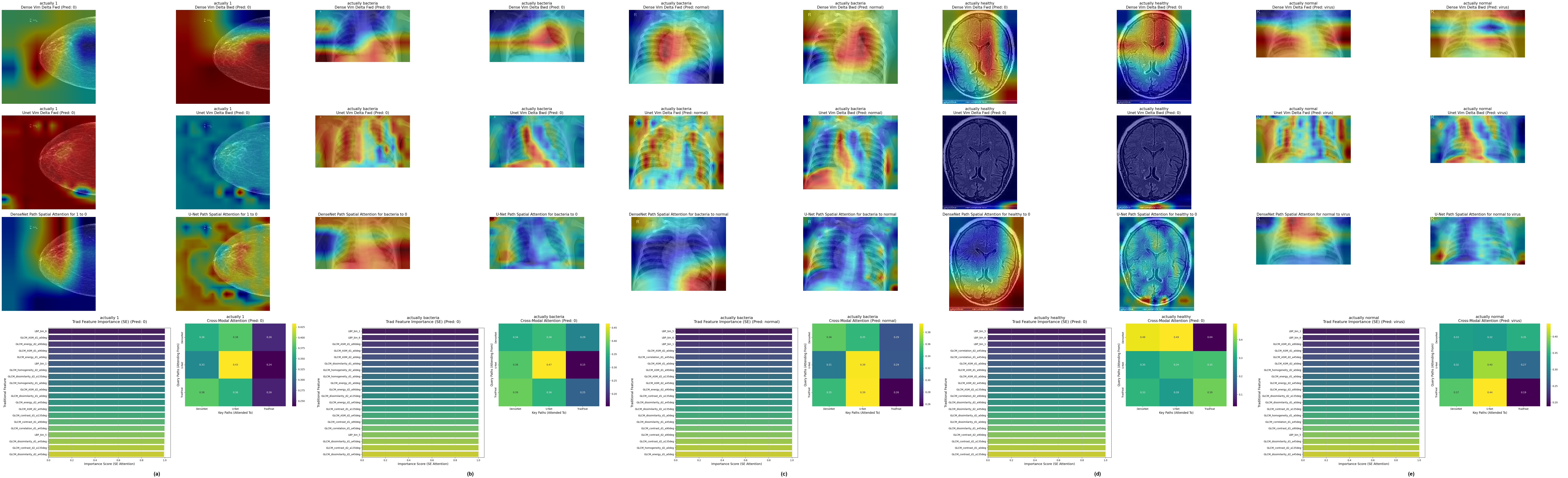}
    \caption{Explainable AI (XAI) Analysis of Representative Misclassified Test Samples by EVM-Fusion. This figure presents a consolidated view of intrinsic explainability outputs for five misclassified samples from the 9-class test set. Noted from (a) to (e). Each sample is analyzed using eight distinct XAI visualizations. From top to bottom, from left to right, they are DenseNet Vision Mamba Forward $\Delta$ Map, DenseNet Vision Mamba Backward $\Delta$ Map, U-Net Vision Mamba Forward $\Delta$ Map, U-Net Vision Mamba Backward $\Delta$ Map,  DenseNet Path Spatial Attention,  U-Net Path Spatial Attention, Traditional Feature Importance (SE Attention bar plot), Cross-Modal Attention Weights (heatmap). These visualizations aim to elucidate the model's internal feature prioritization and fusion dynamics in instances of incorrect classification, providing insights into potential failure modes and areas of confusion.} 
    \label{Fig.2}
\end{figure*}

For (a) in Fig~\ref{Fig.2} both DenseNet and U-Net only focus on the same area of the breast. The model judges the malignant tumor (1) as benign (0), which indicates that there may be no tumor in this area. For (b) and (d) in Fig~\ref{Fig.2}, it can be found that the DenseNet path focuses on the lower right corner, and U-Net pays almost no attention to the human body. This causes the model to even make cross-category errors, mistaking photos of the brain and chest for breasts. And (c) and (e) in Fig~\ref{Fig.2} also make the same mistake. Neither DenseNet nor U-Net pays good attention to the key organ - the lungs.

Through the above analysis, the clinical application of the model can be enhanced. By analyzing these heat maps, if the model's paths do not pay enough attention to the potential lesion area, even doctors who know nothing about the model's principle can know that the output is unreliable. On the contrary, if both models pay attention to the organs that need attention, especially the U-Net path, then even if the doctor does not believe the output of the model, he can focus on the places that U-Net pays attention to and overlaps with the organs, because those places are likely to be the locations of the lesions.

\section{Conclusion}
\label{sec:conclusion}
This paper introduced EVM-Fusion. It is a synergistic framework designed for the clinical task of rapid, holistic medical image triage. We started by identifying the specific challenges of this task. These challenges include the need to process both local and global information and the difficulty of fusing heterogeneous features. We then presented a solution based on a Mamba-enhanced multi-path design and a learnable algorithmic fusion strategy.

Our experiments provided strong validation for this approach. The extensive ablation studies showed that every component of our framework is essential. Their combination leads to performance that is greater than the sum of the parts. Our model also outperformed a wide range of powerful, single-paradigm SOTA models. This shows the practical benefits of our synergistic design. Finally, the built-in XAI tools offer a necessary layer of transparency, which is crucial for building trustworthy AI solutions for clinical use.

In conclusion, EVM-Fusion offers a powerful, interpretable, and generalizable framework for medical image classification. By combining cutting-edge representation learning with a novel algorithmic fusion strategy and a strong emphasis on explainability, this work paves the way for more trustworthy and effective AI-driven solutions in medical diagnostics.

\section*{References}

\bibliography{references}
\bibliographystyle{IEEEtran}

\begin{IEEEbiographynophoto}{Zichuan Yang} photograph and biography not available at the
time of publication.
\end{IEEEbiographynophoto}

\begin{IEEEbiographynophoto}{Yongzhi Wang} photograph and biography not available at the
time of publication.
\end{IEEEbiographynophoto}

\end{document}